\documentclass{sigchi}%{acmart}
\usepackage{booktabs} % For formal tables

\usepackage{algorithm}
\usepackage{algorithmic}

\usepackage{balance}       % to better equalize the last page
\usepackage{graphics}      % for EPS, load graphicx instead 
\usepackage[T1]{fontenc}   % for umlauts and other diaeresis
\usepackage{txfonts}
\usepackage{mathptmx}
\usepackage[pdflang={en-US},pdftex]{hyperref}
\usepackage{color}
\usepackage{booktabs}
\usepackage{textcomp}

\usepackage{microtype}        % Improved Tracking and Kerning
\usepackage{ccicons}          % Cite your images correctly!
\usepackage{todonotes}

\def\plaintitle{Beyond Winning and Losing}

\def\emptyauthor{}
\def\plainkeywords{Authors' choice; of terms; separated; by
  semicolons; include commas, within terms only; required.}

\def\pprw{8.5in}
\def\pprh{11in}

\definecolor{linkColor}{RGB}{6,125,233}
\hypersetup{%
  pdftitle={\plaintitle},
  pdfauthor={\emptyauthor},
  pdfkeywords={\plainkeywords},
  pdfdisplaydoctitle=true, % For Accessibility
  bookmarksnumbered,
  pdfstartview={FitH},
  colorlinks,
  citecolor=black,
  filecolor=black,
  linkcolor=black,
  urlcolor=linkColor,
  breaklinks=true,
  hypertexnames=false
}

\makeatletter
\def\@copyrightspace{\relax}
\makeatother

\begin{document}
\title{Beyond Winning and Losing: Modeling Human Motivations and Behaviors Using Inverse Reinforcement Learning}

\numberofauthors{3}
\author{%
  \alignauthor{Baoxiang Wang\\
    \email{bxwang@cse.cuhk.edu.hk}}\\
  \alignauthor{Tongfang Sun\\
    \email{tongfs@uw.edu}}\\
  \alignauthor{Xianjun Sam Zheng\\
    \email{sam.zheng@gmail.com}}\\
}

\maketitle

\begin{abstract}

In recent years, reinforcement learning (RL) methods have been applied to model gameplay with great success, achieving super-human performance in various environments, such as Atari, Go, and Poker.
However, those studies mostly focus on winning the game and have largely ignored the rich and complex human motivations, which are essential for understanding different players' diverse behaviors.
In this paper, we present a novel method called Multi-Motivation Behavior Modeling (MMBM) that takes the multifaceted human motivations into consideration and models the underlying value structure of the players using inverse RL.
Our approach does not require the access to the dynamic of the system, making it feasible to model complex interactive environments such as massively multiplayer online games.
MMBM is tested on the World of Warcraft Avatar History dataset, which recorded over 70,000 users' gameplay spanning three years period.
Our model reveals the significant difference of value structures among different player groups. 
Using the results of motivation modeling, we also predict and explain their diverse gameplay behaviors and provide a quantitative assessment of how the redesign of the game environment impacts players' behaviors.

%Intelligent user interface research, which investigates interactions between users and computer agents, has become one of the most popular topics in HCI field in recent years. 
%However, previous studies are mostly descriptive, which focus on analyzing the behavior of the users instead of modeling the underlying motivation.
%Their approaches, which commonly involve statistical methods, would incur substantial bias under complex and dynamic UI scenarios.
%Inspired by the recent success of inverse reinforcement learning (IRL) algorithm, we present a novel method to model the underlying motivations of the user behavior instead of the behavior itself.
%The IRL-based method is significantly more powerful than those based on statistics, and can capture the essential user motivations even when the motivations are complex.
%We tested our method on the game of World of Warcraft (WoW), which is a massive online computer game involving different types of users and their respective perspectives and motivations.
%Across over 70,000 users' data spanning three years period, we present our model of user motivation and subsequently the predicting power on user behavior.
%Our research on user motivation is also the base of further UI research, such as analyzing the diverse motivations between different group of users, and the impact of redesigning of the computer agents on the user behavior. 
\end{abstract}

%\keywords{Reinforcement learning, inverse reinforcement learning, player motivation, behavior analysis}

\maketitle

\section{Introduction}

In recent years, reinforcement learning (RL) methods have been applied to model gameplay with great success, achieving super-human performance in various environments, such as Atari, Go, and Texas hold'em poker \cite{mnih2015human,silver2017mastering,moravvcik2017deepstack}.
Those studies, however, primarily focus on winning the game, and the goal of the computer agent is to take actions that can maximize the cumulative scalar rewards, such as achieving high scores or beating the opponents. They have mostly ignored the rich and complex human motivations, which are essential for understanding different players' reward mechanism as well as their complex and diverse behaviors. 
In fact, numerous behavioral and psychology studies \cite{schultheiss2007long,alvarado2005playing} have shown that when people are playing games, apart from competing and winning, they also try to connect with others, or they just want to have some fun or enjoyment by themselves. An extensive survey of game motivation \cite{yee2006motivations,yee2006demographics} with 30,000 players on Massively-Multiplayer Online Games (MMOGs) confirms that human players have complex and multifaceted motivations.
As shown in Tbl. \ref{tbl:components}, the study categorizes the complex motivations of gameplay into ten different types and three different groups, namely, Achievement, Social, and Immersion. 
\begin{figure}[t]
    \centering
    \includegraphics[width=0.5\textwidth]{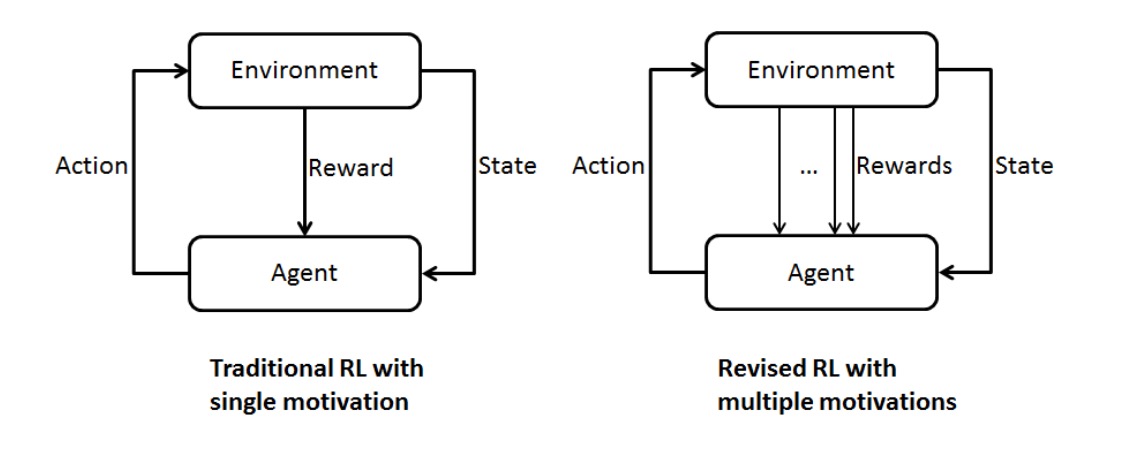}
    \caption{
    In the typical RL model (left), an agent or player has only one single motivation and maximizes one scalar reward. In MMBM (right), an agent or player has multiple motives and the goal is to optimize the combination of different rewards based on each agent's value structure.}
    \label{fig:chi}
\end{figure}

In this paper, we propose a novel method called Multi-Motivation Behavior Modeling (MMBM) that is based on RL and takes into consideration the multifaceted human motivations. 
The objective of MMBM is to model the underlying value structure of the players from the observed human behavior. 
By incorporating the motivation theory in \cite{yee2006motivations} of gameplay, we extend the standard RL framework to cover multiple rewards situations. 
In MMBM, the goal of the agents (or players) is not simply to maximize one scalar reward under one single motivation such as achieving high scores, but instead to maximize the combination of multiple rewards based on the multi-faceted motivations. 
Fig. \ref{fig:chi} illustrates the difference between the typical RL vs. our proposed MMBM. 
The challenge in discovering human' motivations is that they are not explicitly observable. 
Instead, we have to infer them from the players' behaviors, which can be achieved by using inverse reinforcement learning (IRL). 
In MMBM, we extent IRL to uncover the complex, multi-dimensional reward mechanism. 
Our model first quantifies each dimension of the reward signal individually Based on the motivation theory.
The individual signals are subsequently combined under the assumption that each player appears in the trajectory are acting at the best of optimizing their objectives. 
In this way, decomposition of the full reward signal is reduced into a linear program, which is solved efficiently, and subsequently the value structures of the players can be computed.

\begin{table}[t]
\label{tbl:game-motivation}
\noindent
\caption{Components of game motivation}
\vskip 0.15 in
\begin{tabular}{p{2cm} p{6cm}}
    \toprule
    Components & Sub-components \\
    \midrule
    Achievement & Advancement, Mechanics, Competition \\
    Social & Socializing, Relationship, Teamwork \\
    Immersion & Discovery, Role Playing, Customization, Escapism \\
    \bottomrule
    \label{tbl:components}
\end{tabular}
\vskip -0.25 in
\end{table}

A significant advantage of MMBM is that it utilizes only off-policy learning: Each of the individual reward signals is estimated by Q-learning with deep Q-networks (DQN), and MMBM's IRL algorithm takes only the trajectories as its input. 
In this way, MMBM does not require a simulator of the environment, nor does it inquire human players' counterfactual actions that do not exist in the dataset. 
This is beneficial since most of the existing IRL methods have to have access to either the simulation environment or actual human policies, which are usually costly to obtain or simply do not exist. 
For large and complex games, MMBM provides a feasible way to analyze the historical data.

We apply MMBM to model the players' behaviors and motivations of World of Warcraft, which is one of the most successful massively multiplayer online role-playing games with millions of subscribers worldwide. 
We test the MMBM on the World of Warcraft Avatar History (WoWAH) dataset \cite{lee2011world} with 70,000 users' gameplay spanning over a three-year period. 
Our method outputs the value structure which is the most succinct description of the game environment in the perspective of the human players. 
On top of the value structure, it also predicts the players' behaviors accurately, outperforming existing approaches such as large-margin Q-learning \cite{piot2013learning} and policy imitation via classifier. 
Moreover, it reveals the different reward functions and diverse value structures among different player groups, which interestingly agrees with previous knowledge-based studies on WoW \cite{ducheneaut2006alone,nardi2006strangers}

\section{Preliminaries}

\subsection{Inverse Reinforcement Learning}

The process of recovering the reward function from observed trajectories is inverse reinforcement learning (IRL).
It reverses the input and output pairs of RL algorithms, computing the rewards function according to the policies or actions of the agents.
The basic assumption of IRL is that, though the reward function is unknown, it exists and the agents' actions are conducted to maximize the cumulative reward.
The assumption is illustrated by a few different forms in mathematic forms, including linear IRL \cite{abbeel2004apprenticeship,ng2000algorithms}, max-entropy IRL \cite{ratliff2006maximum}, and large-margin Q-learning \cite{piot2013learning}, and etc.
Most of the existing IRL algorithms have either of the two requirements: they need to access the dynamics of the environment \cite{abbeel2004apprenticeship,mnih2015human}, which is usually provided by a simulator of the game; or to access the policy function \cite{zinkevich2007regret}, which requires the agent to retroactively compute the counterfactual action at a historical decision point.
Such requirements are expensive in complex and massive games where human players involved. Hence, an IRL algorithm without those need is desired.
Approaches such as large-margin Q-learning \cite{parameswaran2010large,hester2018deep} and our proposed MMBM do not require that and are suitable for complex, real-world game environments.

\subsection{Deep Q-Learning and Large Margin Q-Learning}

We first define some notions in RL.
In a game environment, at each round $t$, the player conducts an action $a_t$ according to their own policy $\pi(\cdot)$ and current game state $s_t$.
The player may not be able to obtain the full game state (such as events happened out of the vision of the player) and uses the observation $x_t$ to substitute $s_t$.
The player subsequently receives the feedback from the environment, including a scalar reward $r_t$ and the observation $x_{t+1}$ of the next round.
The player's intention is to maximize his/her discounted cumulative reward, also known as the action-value function, 
\begin{equation}
Q^\pi (s,a)=\mathbb{E}[R_t | s_{t}=s, a_{t}=a, \pi],
\end{equation}
where $R_t=\sum_{t^\prime\geq t}\gamma^{t^\prime-t}r_{t^\prime}$.
Deep Q-Learning uses a Deep Q-Network (DQN) to estimates the $Q^{\pi^\ast}(s,a)$ value, where $\pi^\ast$ is the maxima of the Q-value over all policies.
DQN uses the recursive relation of $Q^{\pi^\ast}(s,a)$, known as the Bellman equation
\begin{equation}
Q^{\pi^\ast}(s_t,a_t)=r_t + \gamma \max_{a^\prime}Q^{\pi^\ast}(s_{t+1}, a^\prime). \label{eqn:bellman}
\end{equation}
The estimation is conducted by minimizing the Bellman error
\begin{equation}
L_1=\mathbb{E} [\frac{1}{2}(Q^{\pi^\ast}(s_t,a_t|\theta) - y^\prime )^2],
\end{equation}
where $y^\prime = r_t - \gamma\max_{a^\prime}Q^{\pi^\ast}(s_{t+1}, a^\prime|\theta^\prime)$ is the target value function and $\theta^\prime$ the parameter of the target network. 

While DQN estimates the $Q(s,a)$ function of $\pi^\ast$ from the reward signals, which is subsequently used to retrieve the optimal policy, large-margin Q-learning approximates the action-value function corresponding to the observed behavior directly.
Suppose the policy and the reward signals of the player or a group of players are unknown and we have observed a set of state-action pairs generated by such a policy.
Large-margin Q-learning \cite{parameswaran2010large,hester2018deep} assumes (as most of the IRL algorithms do) that the players' actions are intended to maximize their action-value, namely,
\begin{equation}
Q^\ast(s,a) \geq \max_{a^\prime \in \mathcal{A}(s)}Q^\ast(s,a^\prime)
\label{eqn:irl}
\end{equation}
is satisfied for all state-action pairs with a margin.
Note that $\mathcal{A}(s)$ is the set of all feasible actions under state $s$.
Adding a large margin toward the difference between inequality \eqref{eqn:irl}, it results in the error term
\begin{equation}
L_2 = \frac{1}{2}(Q^{\pi^\ast}(s_t,a_t) - (l_{s,a}+\max_{a^\prime}Q^{\pi^\ast}(s_t,a^\prime))^2,
\end{equation}
where $l_{s,a}$ is the margins, which could either be pre-defined parameters trainable parameters.

\section{Methods}

\subsection{Reward Mechanism Modeling in MMBM}

\begin{figure*}[t]
  \centering
  \includegraphics[width=\textwidth]{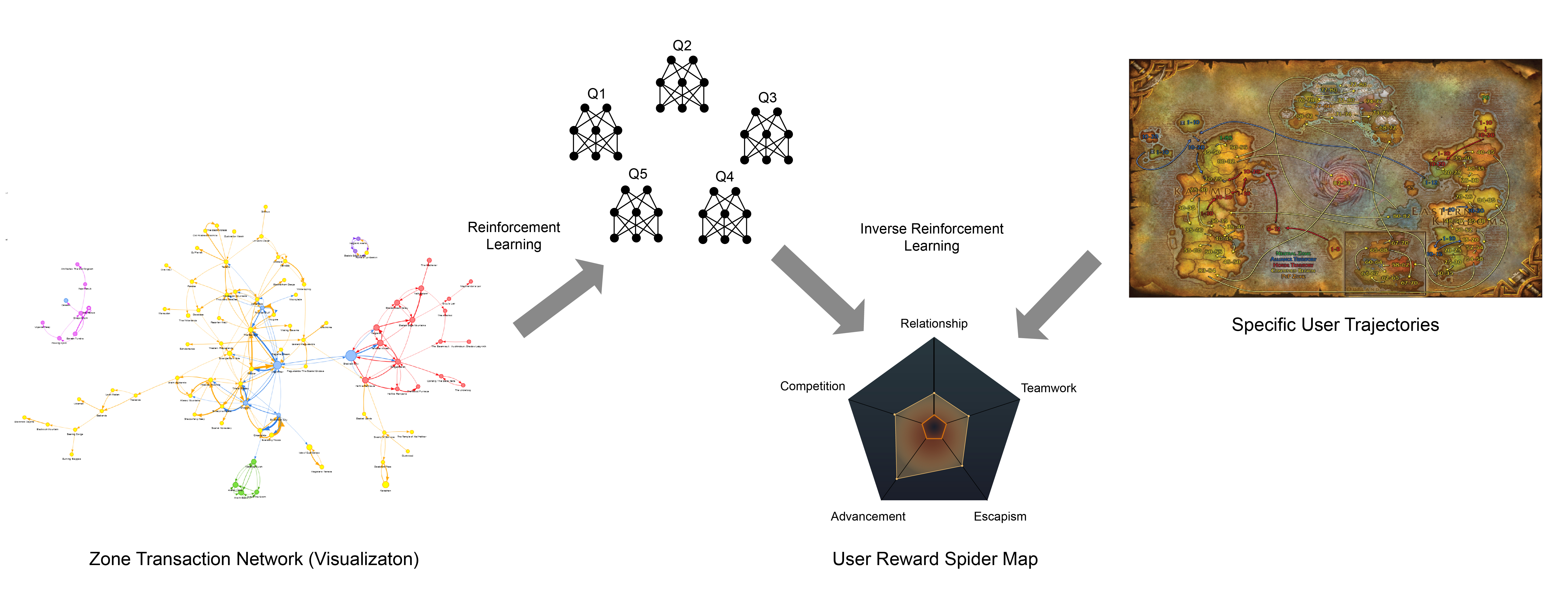}
  \caption{Illustrative execution of Alg. \ref{alg:mmbm} on WoWAH}
  \label{fig:workflow}
\end{figure*}

We present our MMBM algorithm to compute the underlying reward function of the agents. 
Our method can be viewed as a two-step workflow: The first step, known as Q-learning, estimates the reward functions of the players at different states of gameplay environment; The second step, a variant of inverse reinforcement learning (IRL), estimates the combination or weights of the different rewards learned at the first step.
In essence, the two-step methodology decomposes the complex interactions between players and a game environment into multiple quantitative metrics and solve them separately. 
An intuitive illustration of the two-step framework on WoWAH is shown in Fig. \ref{fig:workflow}, while the formal algorithm is described in \ref{alg:mmbm}.

The fundamental idea behind the first step is that in a complex environment, given the same situation or state, different players respectively perform their optimal actions and exhibit diverse behaviors.  For example, a player who values more about the relationship with his/her teammates would spend more time on team-based activities than those players who focus more on their advancements or achievements.
Because he/she receives more overall rewards, by getting more teamwork-based rewards that he/she values.
Hence, the combination of multiple reward signals is essential to model the users' behavior and their underlying value structure.

MMBM learns the weights of the combination of multiple motivations from the user behavior data using IRL techniques.
Formally, let $\mathcal{T}$ be the set of state-action pairs of a user or a group of users.
It consists of the choices of the users (correspond to the term \textit{actions $a_t$} in RL; $t$ stands for time step index) under various of situation or scenarios (correspond to the term \textit{states $s_t$} in RL).
We can also infer from the states that the agents are receiving feedback on multiple rewards  $f_t=(f_t^1,\dots, f_t^n)$ simultaneously.
Assume that the players are optimal in processing any information available to them and display optimal trajectory towards their objectives \footnote{The assumption is reasonable as we take multiple dimension of reward signals into consideration}. Then, given the same environment state, different players perform diverse actions or display complex behaviors must be resulted from their different motivations or value structure  $f_t^1,\dots, f_t^n$. With the assumption, it reduces to find a valid combination of rewards such that under that combination every action is optimal, that is, there does not exist another feasible action that yields a higher total reward.
Let $\phi \in \mathbb{R}^n$ be the combination weights, subjecting to $||\phi||_1=1, \; \phi \geq 0 $, define the reward as 
\begin{equation}
r_t = \phi^Tf_t
\end{equation}
The action-value function (or Q-function) describes the objective of the user
\begin{equation}
Q^\ast(s,a)=\mathbb{E}[R_t | s_{t}=s, a_{t}=a, \pi^\ast],
\label{eqn:q}
\end{equation}
where
\begin{equation*}
R_t=\sum_{t^\prime\geq t}\gamma^{t^\prime-t}r_{t^\prime}.
\end{equation*}
Q function gives the expected cumulative rewards the user gets if the user chooses optimal action $a$ under state $s$ and follows the best policy thereafter.
It is the function that should satisfy the previously discussed action optimality, which can be formulated as 
\begin{equation}
\label{eqn:prime}
Q^\ast(s,a) \geq \max_{a^\prime \in \mathcal{A}(s)}Q^\ast(s,a^\prime),
\end{equation}
where $a^\prime \in \mathcal{A}(s)$ is the set of all possible actions the user can take at the state $s$.
Since $Q^\ast(s,a)$ is a function of $\phi$, solving inequation \eqref{eqn:prime} will yield the combination weight $\phi$ we want.

Though $\eqref{eqn:prime}$ itself could be infeasible and hard to solve, we apply two approximation to find the solution.
First, let $Q^i(s,a)$ be the action-value function, as if $f^i$ is the only existing reward signal
\begin{equation}
Q^i(s,a)=\mathbb{E}[\sum_{t^\prime\geq t}\gamma^{t^\prime-t}f^i_{t^\prime} | s_{t}=s, a_{t}=a, \pi^{i}] \label{eqn:qi}
\end{equation}
At this moment we assume the such $Q^i$ function can be accurately estimated. 
We then apply linear scalarization \cite{piot2013learning} from IRL to explicitly separate out the weights $\phi$
\begin{equation}
\label{eqn:scalarization}
Q^*(s,a) = \phi^T\tilde{Q}(s,a),
\end{equation}
where the vector of function $\tilde{Q}(s,a)=(Q^1(\cdot),\dots,Q^n(\cdot))$.
Second,  we introduce the slack variables $\xi_{s,a}$, which models the cases that users behave less optimally, such as making mistakes or just playing randomly.
$\xi_{s,a}$ sets the threshold of the difference between the actual action-value $Q^\ast(s,a)$ and the largest possible action-value $\max_{a^\prime \in A(s)}Q^\ast(s,a^\prime)$ over all feasible actions.
The value of $\xi_{s,a}$ is positive whenever inequality \eqref{eqn:prime} is not satisfied, and zero otherwise.
Solving the inequation \eqref{eqn:prime} is reduced to minimizing the summation of the slack variable $\xi_{s,a}$ over all observed pairs, which is
\begin{equation}
-\sum_{s,a} \left[\min(0, Q^\ast(s,a) - \max_{a^\prime \in A(s)\backslash a}Q^\ast(s,a^\prime))\right]. \label{eqn:slack}.
\end{equation}
After the two approximation steps, minimizing such total slacks is reduced into a linear program (LP) problem as follows.

Given that the action-value function $\tilde{Q}(s,a)$ for each of the reward signal, and let $\mathcal{T}$ be the set of observed state-action pairs of (that is, our dataset), minimization of the summation of the slack variables \eqref{eqn:slack} is formulated into the following LP
\begin{equation}
\begin{aligned}
& \underset{\phi, \xi}{\mathrm{minimize}}
& & \sum_{s,a} \xi_{s,a} \\
& \mathrm{subject}\text{ }\mathrm{to}
& & \phi^T(\tilde{Q}(s,a)-\tilde{Q}(s,a^\prime)) \geq - \xi_{s,a}, \\
&&& \quad\quad\quad \forall (s,a) \in \mathcal{T}, a^\prime \in \mathcal{A}(s)\backslash a \\
&&& \phi \geq 0, \; ||\phi||_1\geq 1\\
&&& \xi_{s,a} \geq 0, \; \forall (s,a) \in \mathcal{T}. 
\label{eqn:lp}
\end{aligned}
\end{equation}
As LP can be solved efficiently, MMBM finds the composition of the rewards by solving the weights $\phi$ of different reward signals in Eq. \eqref{eqn:lp}.

The remaining problem is to estimate the action-value action-value function $\tilde{Q}(s,a)$, which is solved by Q-learning via DQN.
Referring to Algorithm \ref{alg:mmbm}, line \#13-16 are the decomposing part which is reduced to LP \eqref{eqn:lp} and line \#4-12 are the DQN approach in MMBM, which is standard in RL and detailed in the next section with respect to our environment settings.
With our two-step approach, MMBM takes the history of state-action pairs as input, which is usually logged during the gameplay.
It solves $\phi$ , which is a quantitative description of human players' motivations and the value structure.
\begin{algorithm}[tb]
   \caption{MMBM}
   \label{alg:mmbm}
\begin{algorithmic}[1]
   \STATE {\bfseries Parameters:} learning rate $\alpha$, discount factor $\gamma$
   \STATE {\bfseries Initialization:} initialize network parameters $w^i$ randomly
   \STATE {\bfseries Input:} set $\mathcal{T}$ of trajectories
   \FOR{$i=1$ {\bfseries to} $n$}
     \FOR{$t$ {\bfseries to} size of $\mathcal{T}$}
     \STATE Calculate $f_t^i$
     \ENDFOR
     \REPEAT 
       \STATE Compute $L_1^i=\mathbb{E} [\frac{1}{2}(Q^i(s_t,a_t|w^i) - y^\prime )^2]$
       \STATE Update $w^i = w^i - \alpha \nabla_{w^i} L_1^i$
     \UNTIL {convergence of $Q^i(s,a)$}
   \ENDFOR
   \FOR{$t$ {\bfseries to} size of $\mathcal{T}$}
     \STATE Compute $\tilde{Q}(s,a)=(Q^1(\cdot),\dots,Q^n(\cdot))$
   \ENDFOR
   \STATE Find $\phi$ by solving linear program \eqref{eqn:lp}
\end{algorithmic}
\end{algorithm}

\subsection{Off-Policy Action-value Function Approximation}

\begin{figure*}[t]
  \centering
  \includegraphics[width=\textwidth]{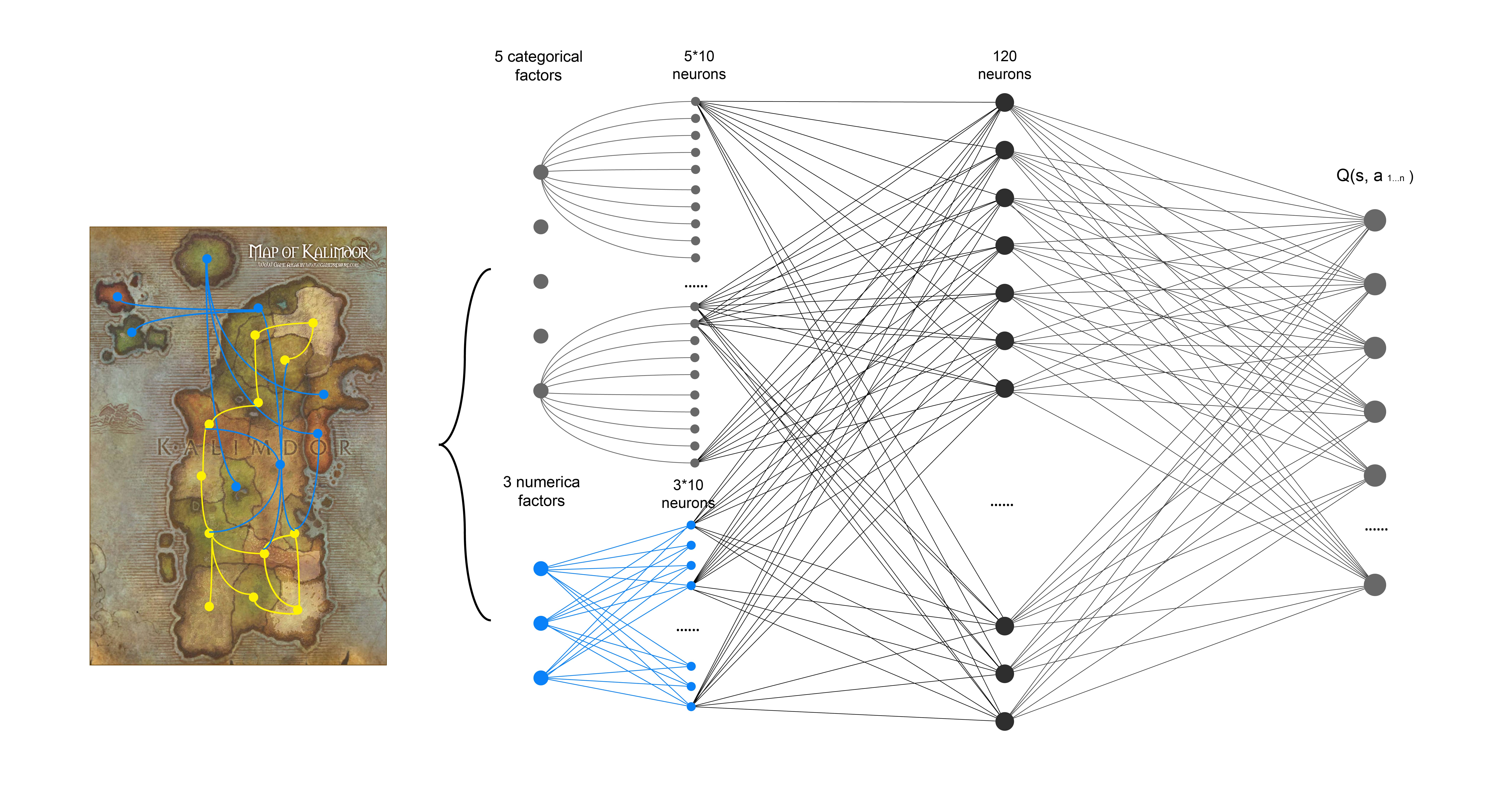}
  \caption{DQN architecture for $Q^i$ training, $i=1, \dots,n$, on WoWAH}
  \label{fig:architecture}
\end{figure*}
In Alg. \ref{alg:mmbm}, MMBM requires the approximation of the action-value function $Q^i(s,a)$ for each of the component of the rewards .
Such approximation should be a fair estimation of the cumulative reward the user would receive if the user chooses an action $a$ at the state $s$ and maximizes the $i$-th reward thereafter.
DQN uses the recursive property (i.e. Bellman equation) that the action-value estimator should have, that is, the cumulative reward since the current step onwards should be the immediate reward plus the cumulative reward since the next step onwards.
Using the property, DQN updates the action-value function iteratively, by moving the $Q^i(s,a)$ value toward (by upgrading the parameters) it's target $r_t + \gamma Q^i(s^\prime,a^\ast)$.
By the time $Q^i(\cdot)$ converges to satisfy the Bellman equation, it estimates the action-value given any state-action pair.

An advantage of Q-learning is that it learns off-policy property, which implies that the action-value function approximation does not rely on real-time data or any game simulator.
To understand this, we observe that the $(s,a,s^\prime,a^\ast$ tuples used in the iterations of the update could be feed into the model with an arbitrary order and could involve any $a$ without being required that $a$ is generated by a certain policy.
It is very important to our algorithm because since the interaction records between the agent and the environment, such as the computer-human iteration, are usually available in its offline mode.
This means our MMBM does not require the dynamics of the environment for training the model. 
Using gameplay historical data or player behavior log data as the input, MMBM models complex game environments such as massively multiplayer online games.

The function approximator which parametrizes the action-value functions largely depends on the environment.
Taking our experiments on the WoWAH dataset as an example, the DQN architecture is designed according to the available observations and is applied to all reward signals $i=1,\dots,n$.
As shown in Fig. \ref{fig:architecture}, the categorical elements of the input (e.g. \textit{race}, \textit{class}, etc.) are first processed by an embedding layer \cite{mikolov2013distributed}, while the numeral elements (e.g. session length, current level etc.) are first fed into a fully connected (FC) layer with rectifier non-linearity. 
The output of embedding layer and FC layer are then concatenated and fed into another FC layer with rectifier non-linearity. 
A final FC layer is applied to compute the $Q(s,a)$ value for each action $a\in\cup_s\mathcal{A}(s)$.
The detailed introduction of the environment and the details of each of the input variables are included in the experiment section.

\subsection{Imitation Learning and Predictions}

An immediate use of the action-value function is to derive the optimal policy which imitates the gameplay of the samples.
That is, let $\pi^\ast(s)$ denote the action at state $s$
\begin{equation}
\pi^\ast(s) = \text{argmax}_{a^\prime}Q^\ast(s,a^\prime)
\label{eqn:policy}
\end{equation}
is the policy function which predicts the players move.
The intuitive understanding of our predicting power is if the players' reward system is available, we could easily predict the players' behavior.
Moreover, the predictions of the user behavior are with a reason behind: While most of the classification models are just black boxes their outputs may not be corresponding to a clear intuitive. MMBM, instead, reveals the underlying system that drives the behavior before making predictions

To make predictions, MMBM first solving LP \eqref{eqn:lp}, and get the combination of reward signal $r_t$.
As the action-value function $Q^i$ have been already learned, the action-value function   $Q^\ast(s,a)$ becomes known by applying Eq. \eqref{eqn:scalarization}.
To avoid the bias involved in the scalarization process, we learn the $Q^\ast(s,a)$ using the combination weights and the original dataset once more.
The re-train of then action-value function can be generalized to a more complex combination, for example, Pareto combination of individual reward signals.
Note that, $Q^\ast(s,a)$ and $\pi^\ast(s)$ are not corresponding to just advancement or fastest leveling up in the game. 
MMBM is beyond winning and losing: it models and predicts the actual actions that the humans would have conducted once they present at such a state.

\section{Experimental Analysis}
\label{sec:experiments}

\subsection{WoWAH Player Behavior Dataset}

We tested our MMBM in WoW, one of the most successful MMORPGs in the world with millions of active players. Like other MMORPGs, each player chooses a character avatar and control the avatar in third- or first-person view throughout the game. Players can explore the landscape, fight various monsters, complete quests individually or cooperatively,  communicating and interacting with other players, or build their guilds (groups).
As shown by Yee \cite{yee2006motivations,yee2006demographics}, players' motivations are distinct and their actions and behavior are complicated. 
The WoWAH dataset \cite{lee2011world} is an interesting dataset to investigate the behavior. It records a significant amount of gameplay data with over 70,000 players' movements (regarded as actions) from realm \textit{TW-Light's Hope} every 10 minutes spanning for the 3-year period.
Previous studies on this dataset are either based on descriptive statistics  \cite{lee2011world} or using simple classifiers or clustering \cite{suznjevic2011mmorpg,drachen2014comparison,bauckhage2015clustering,Bell2013a}, and these methods fail to capture the rich and complex motivations of the players. 

%\begin{table}[t]
%    \centering
%    \caption{WoWAH Dataset Attributes}
%    \begin{tabularx}{\textwidth}{lX}
%        Attribute & Value \\
%        \midrule
%        Duration & 1107 days \\
%        Sample Interval & Every 10 minutes \\
%        \#Users & 70,000+ \\
%        Locations & One of 165 zones
%        \label{tbl:wowah}
%    \end{tabularx}
%\end{table}

%WoWAH contains 70,055 users from realm \textit{TW-Light's Hope}, each with 5162.1 minutes spent online on average \footnote{We filter out some users with too short playing time}.
%At each mark of the 10-minute interval, the users' locations and other information such as level, guild status, and class are recorded.
From the reinforcement learning perspective, we treat each player as a human agent who conducts an action at each time interval.
All available data such as current level or joining the \textit{guild} are regarded as observations.
The players' trajectories are composed of a sequence of locations and observations, which partially reflect their playing strategies.
\cite{Bell2013a,shen2014characterization}.
Even though Yee theorized ten different motivation for gameplay,  we apply five of them that are frequently observed in the WoWAH dataset. Therefore, we compute five different kinds of motivations using the WoWAH dataset and let $n=5$ in Alg. \ref{alg:mmbm}.
The constructions of the motivation values $f_1,\dots, f_5$ are illustrated in Tbl. \ref{tbl:satisfactions}, and are based on the Yee's research and other WoW case studies \cite{ducheneaut2006alone,nardi2006strangers}.
With those value, we model the final reward function that each player tries to maximize during the gameplay.

\begin{table}[t]
\caption{Different types of motivations in WoWAH and corresponding definitions}
\vskip 0.15 in
\begin{tabular}{p{1cm} p{7cm}}
    \toprule
    Motv. & \textbf{Category} \& Definition \\
    \midrule
    $f^1$ & \textbf{Advancement} describes how fast the player levels up in the game. It's the speed the user levels up, divided by the averaged speed at the entire WoWAH. \\
    $f^2$ & \textbf{Competition} describes if the player joins \textit{Battleground} or \textit{Arena} and competing with human opponents. It equals the number of visits. \\
    $f^3$ & \textbf{Relationship} is linear to the duration that the player has been in the current \textit{guild}. \\
    $f^4$ & \textbf{Teamwork} describes the intention of conducting teamwork, which is the number of recent zones with teamwork features visited. Zones with teamwork features include \textit{Battleground}, \textit{Arena}, \textit{Dungeon}, \textit{Raid}, or a zone controlled by \textit{The Alliance}. \\
    $f^5$ & \textbf{Escapism} is the linear combination of the duration of the recent game session and the number of days the player continuously login to the game recently. \\
    \bottomrule
    \label{tbl:satisfactions}
\end{tabular}
\vskip -0.25 in
\end{table}

\subsection{Player Motivation Modeling}

\begin{figure}[t]
    \centering
    \includegraphics[width=0.5\textwidth]{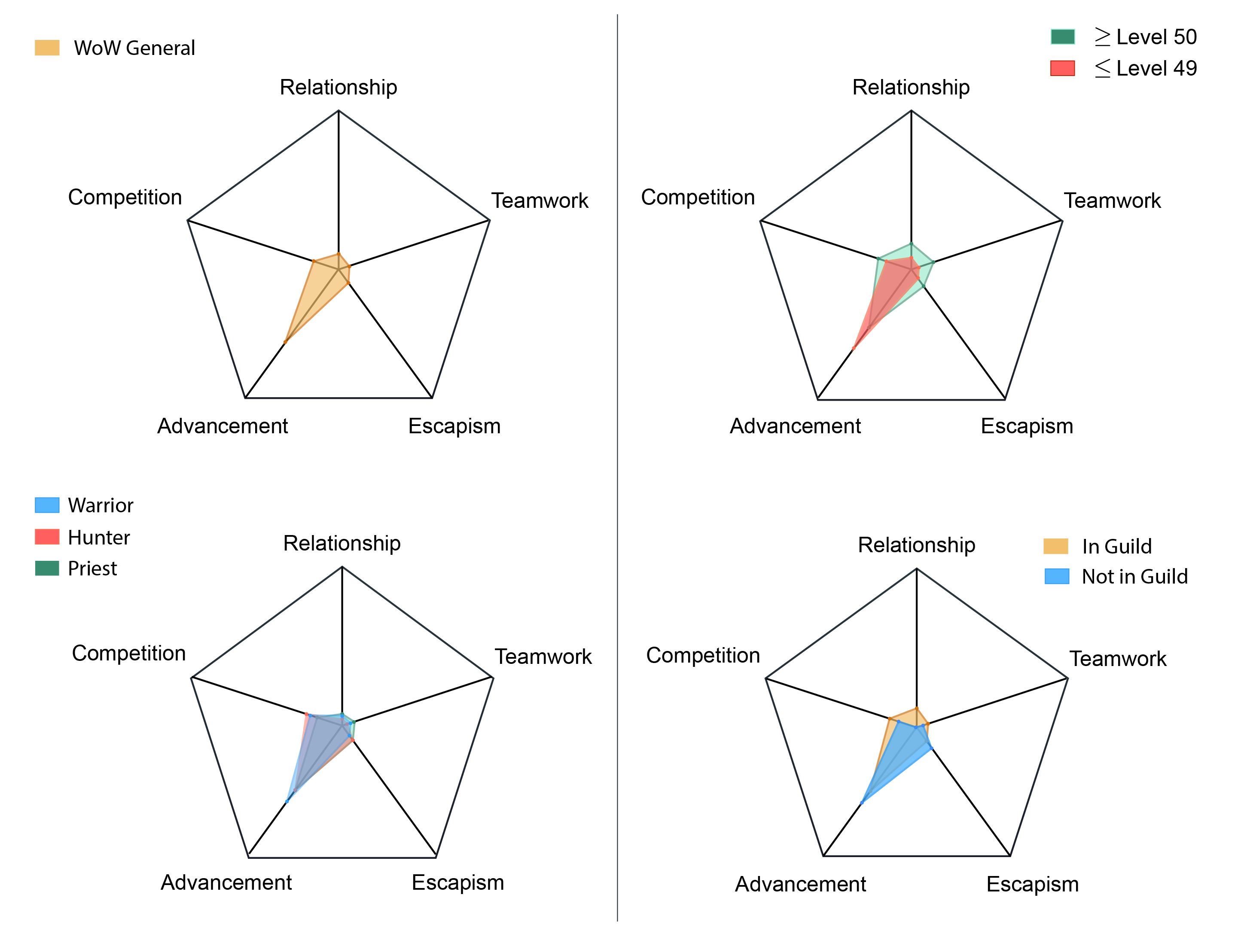}
    \caption{Spider maps to represent player reward mechanism or value structure. \textbf{Top-left:} the weights of different motivations for the entire WoW player community; \textbf{top-right:} different value structures between the players at higher level ($\geq 50$) and the players at lower level ($\leq 49$); \textbf{bottom-left:} different value structures between the players in different classes \textit{Warrior}, \textit{Hunter}, and \textit{Priest}; \textbf{bottom-right:} comparison of different value structure of the players who are in  \textit{guild} and those who are not in a \textit{guild}}
    \label{fig:spider}
\end{figure}

We present our experimental results on recovering the multi-motivation mechanism, which is the solution of LP \eqref{eqn:lp}.
We use Tbl. \ref{tbl:satisfactions} and solving LP \eqref{eqn:lp} on trajectories that are randomly drawn from the WoWAH dataset. The underlying reward mechanism and value structure of the whole player community is 
\begin{equation}
\phi=(0.40,0.10,0.21,0.16,0.12)^T.
\end{equation}
In other words, when choosing an action or conducting a behavior, the players' total motivation is composed of 40\% their advancement, 10\% their competition, 21\% their relationship, 16\% their teamwork, and 12\% their escapism on average.
The results are illustrated as a spider map in Fig. \ref{fig:spider}.
Note that the above $\phi$  is calculated based on the entire player database, and our MMBM can calculate the respective $\phi$ vector for an individual player or a player group.

We show some comparison results for different player groups. Significantly value structures difference is observed between the players at a higher level ($\geq 50$) versus the players at a lower level ($\leq 49$), where the players at the lower level are much more motivated to advance as indicated by the bigger weight on the Advancement motivation. It also shows interesting difference among players in different classes, \textit{Warrior}, \textit{Hunter}, and \textit{Priest}, where the \textit{Warrior} players value more on Advancement and the \textit{Priest} players value more about relationship. It agrees with the common knowledge in WoW that the spells of \textit{Priest} focus on benefiting (healing, buffing, etc.) the team rather than those of \textit{Hunder} and \textit{Warrior} whose spells are more related about damage, and damage/tank, respectively. 
Lastly, the results also show that players in the \textit{guild} value more about Teamwork and Relationship motivations as compared to the players that aren’t in a \textit{guild}. The difference of th weights are distributed into \textit{advancement} and \textit{escapism} instead.
Interestingly, those quantitative results agrees with previous knowledge-based studies on WoW \cite{ducheneaut2006alone,nardi2006strangers}.

\subsection{Predicting Players' Behavior}

Once MMBM models the humans' motivation and value structure, it straightforwardly predicts the complex user behaviors. The prediction is made by the policy stated in Eq. \eqref{eqn:policy}. In WoW, at any given state, the player chooses its movements to stay in the same zone or move to another feasible zone. The action space is discrete, and depending on the players' level the size ranges from a few zones or over one hundred zones. Therefore, the chance of randomly guess players' next action is quite low. Our predictions are quite accurate considering the difficulties and action space size.

We evaluate the accuracy of the prediction, by comparing if the predicted action $\pi^\ast(s)$ agrees with the actual action $a$ for the $(s,a)$ pairs in the dataset.
Experiments show that policies induced by a biased reward function underperform our $\pi^\ast$.
That is computed by adding a disturb factor $\epsilon \sim \mathcal{N}(0,0.05)$ to the solution $\phi$ of LP \eqref{eqn:lp}, as shown in \ref{tbl:accuracy}.
We also compare our result with the policy only focuses on advancement, by setting $\phi=(1,0,0,0,0)^T$, and the results show that our approach predicts players' actions significantly better. This finding indicates that taking into consideration of multiple motivations of the user or player not only can reveal each player's different value structure but can also predict the complex user behavior more accurately. 
Then, we test the large-margin Q-learning and policy imitation and the results show that both these method are less accurate compared with our MMBM.
Note that policy imitation via supervised learning is implemented by a multi-class support vector machine, mapping the state $s$ to the action $a$.

A close examination of the errors that are made during the prediction yields some interesting understandings.
As our MMBM model assumes that every player tries to maximize their cumulative reward in Eq. \eqref{eqn:irl}, i.e.,  everyone is regarded as a rational and optimal player.
Unfortunately,  our model would have some trouble to distinguish whether a particular action that deviates from an average one is caused by the player's actual intention or the player's sub-optimality during the gameplay. For instance, some of the players could spend hundreds of hours on solo \textit{quest} but fail to level up quickly. This could be due to their intention of enjoying doing the quest repeatedly or the players not knowing the optimal strategy to level up. We will address this limitation of our method in the future work by considering humans' different ability or skill levels.

\subsection{Dynamics of the Human Motivation}

\begin{table}[t]
\caption{Accuracy of different approaches}
\vskip 0.15 in
\begin{tabular}{p{1.7cm} p{1.5cm} p{4.2cm}}
    \toprule
    Approach & Accuracy & Notes\\
    \midrule
    $\pi^\ast$ & 56.5\% & From Eq. \eqref{eqn:policy}\\
    Dist. $\pi^\ast$ & 52.5\% & Use $\phi+\epsilon$ instead\\
    $\pi^1$ & 45.9\% & Use $\phi=(1,0,0,0,0)^T$\\
    LMQL & 47.2\% & Large margin Q-learning \\
    PI & 31.0\% & Policy imitation via SL \\
    Linear Q & 29.5\% & Replace DQN w/ linear \\
    \bottomrule
    \label{tbl:accuracy}
\end{tabular}
\vskip -0.25 in
\end{table}

\begin{figure}[t]
    \centering
    \includegraphics[width=0.5\textwidth]{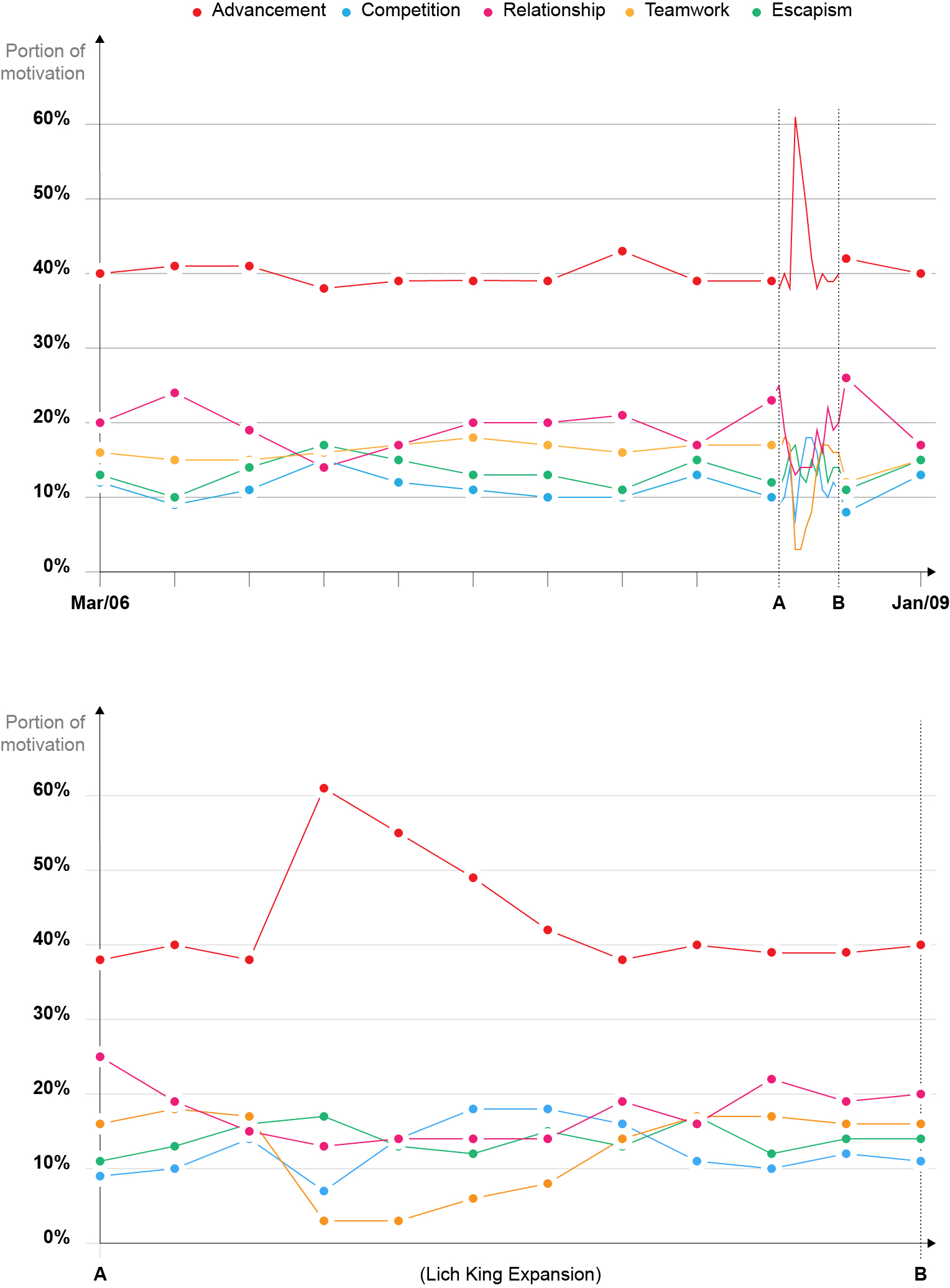}
    \caption{\textbf{Top}: trends of different kinds of motivations during from Mar 2006 to Jan 2009; \textbf{Bottom}: the enlargement of the top figure during around the release of patch \textit{Wraith of the Lich King}}
    \label{fig:trend}
\end{figure}

The motivation of gameplay may evolve. It can also be impacted by the new design or new versions of the game environment. 
How would a design update affect the users' motivations and behaviors and how we can quantify this impact? It's a very interesting question for every game designer to consider.
We conduct an analysis of the dynamics of player motivations on the WoWAH dataset, i.e. how the underlying reward mechanism for players changes over time.
To achieve this, the idea is that the set $\mathcal{T}$ in LP \eqref{eqn:lp} may contain any numbers of trajectories.
Randomly drawing $(s_t,a_t)$ from the dataset where $t$ is restricted to a specific time range yields the set $\mathcal{T}$ which illustrates the player's motivations during that time range.
Taking the time range chronologically, we show the evolution of game motivation, characterized by the elements in $\phi$.
Fig. \ref{fig:trend} illustrates the trend of \textit{Advancement}, \textit{Competition}, \textit{Relationship}, \textit{Teamwork}, and \textit{Escapism}.\footnote{Note that at any time the weights of those elements sum to 1, representing how players value those satisfactions relatively.}.

First, we observed the dramatic increase in Advancement and Competition during the mid-to-late period on the graph.
It happens at around the 150000th time interval, which coincides with the release of the patch \textit{Wraith of the Lich King} on November 2008.
Analyzing the game update patch, two primary reasons can explain the increased level of motivation on Advancement.
First, the patch increased the maximum player level from 70 to 80. As a result, the players with level 70, the previous max level, were rushing to complete the remaining ten leveling ups to reach the new max level. 
Second, the patch introduced two new classes in the game, namely \textit{Death Knight} and \textit{Shaman}, and this gave incentives to many players to open the secondary accounts and to level up them is the first thing to do afterward.
Meanwhile, the reason for more Competition is that many players tend to join player-versus-player (PvP) to compete with other human players to get more familiar with the mechanism of their new avatar.
It's also noticed that the satisfactions are not independent of each other: players spend more time on advancement usually have insufficient time to complete tasks which require teamwork but provides no experience for leveling up.
That's shown in Fig. \ref{fig:trend} that the weight for teamwork decreases each time the weight for advancement increases, and vice versa.

We analysis the overall trend of the game during the three years when WoWAH was collected.
It turns out that the game emphasis more on teamwork and relationship during the period, partially because the dataset was collected only two years after the game release, and the players are getting more and more involved in the game during that time.
Apart from that, the weights of different kinds of motivations are under influences from both game patches and updates, and game user community.
Overall, our MMBM model and analysis provides useful insights in Fig. \ref{fig:trend}  for game designers and researchers.

\section{Conclusions and Future Work}

We present MMBM, a general RL model that takes multifaceted human motivations into consideration.
MMBM conducts the IRL task, while not relying on the access of policy function nor the dynamics of the environment. Hence, MMBM can be applied to study complex, interactive environments with its historical dataset. 
Our experiment results on the WoWAH dataset shows that MMBM recovers reasonable reward mechanism of the players. On top of that, it predicts human players’ behaviors accurately, shows how different group of players have respective value structure, and provides a quantitative assessment of how the redesign of the game environment impacts players' behaviors.

We view our work as one of the first that can combine the richness of psychological and game research theories with the rigorousness of RL models. 
Our goal is beyond winning and losing: not simply to create software agents that beat human in various games or competitions, but to propose methods that can help to understand the intricacy and complexity of human motivations and their behaviors. We hope to inspire more researchers to investigate this topic further.

\bibliographystyle{sigchi}
\bibliography{refs} 

\end{document}